\documentclass[letterpaper]{article} 
\usepackage{aaai25}  
\usepackage{times}  
\usepackage{helvet}  
\usepackage{courier}  
\usepackage[hyphens]{url}  
\usepackage{graphicx} 
\usepackage{natbib}  
\usepackage{caption} 
\usepackage{algorithm}
\usepackage{algorithmic}

\usepackage{algorithm}
\usepackage{algorithmic}
\usepackage{pifont} 
\usepackage{subcaption}
\usepackage{verbatim}
\usepackage{graphicx}
\usepackage{multirow}
\usepackage{xspace}
\usepackage{amsfonts,amssymb} 
\usepackage{microtype}
\usepackage{amsmath}
\usepackage{amsthm}
\usepackage{booktabs}
\usepackage{times}
\usepackage{latexsym}
\usepackage{color}
\usepackage[dvipsnames, table]{xcolor}
\usepackage{soul}
\usepackage{verbatim}
\usepackage{multirow}
\usepackage{xspace}
\usepackage{amsfonts,amssymb} 
\usepackage{microtype}
\usepackage{times}
\usepackage{latexsym}
\usepackage{color}
\usepackage[dvipsnames, table]{xcolor}
\usepackage{soul}
\usepackage{colortbl}
\usepackage{booktabs,makecell, multirow, tabularx}
\usepackage{adjustbox}
\usepackage{xcolor}
\usepackage{listings}

\usepackage{booktabs}
\usepackage{amssymb}
\usepackage{bbding}
\usepackage{pifont}
\usepackage{wasysym}
\usepackage{utfsym}
\usepackage{fontawesome}

\usepackage{mdframed}

\usepackage{pifont}

\usepackage{newfloat}
\usepackage{listings}
\DeclareCaptionStyle{ruled}{labelfont=normalfont,labelsep=colon,strut=off} 
\floatstyle{ruled}
\newfloat{listing}{tb}{lst}{}
\floatname{listing}{Listing}
\title{Revisiting Multimodal Fusion for 3D Anomaly Detection from an Architectural Perspective}
\author{
    Kaifang Long\textsuperscript{\rm 1*},
    Guoyang Xie\textsuperscript{\rm 2}\thanks{Contribute equally.},
    Lianbo Ma\textsuperscript{\rm 1}\thanks{Corresponding author.},
    Jiaqi Liu\textsuperscript{\rm 3},
    Zhichao Lu\textsuperscript{\rm 3}
}
\affiliations{
    \textsuperscript{\rm 1}Software College, Northeastern University, Shenyang, China\\
    \textsuperscript{\rm 2}
    The Department of Intelligent Manufacturing, CATL, Ningde, 
 China\\
    \textsuperscript{\rm 3}The Department of Computer Science, City University of Hong Kong, Hong Kong, China\\

longkf@stumail.neu.edu.cn, guoyang.xie@ieee.org, malb@swc.neu.edu.cn, liu\_jiaqi\_@outlook.com, luzhichaocn@gmail.com
}

\usepackage{bibentry}

\begin{document}

\maketitle

\begin{abstract}
Existing efforts to boost multimodal fusion of 3D anomaly detection (3D-AD) primarily concentrate on devising more effective multimodal fusion strategies. 
However, little attention was devoted to analyzing the role of multimodal fusion architecture (topology) design in contributing to 3D-AD. 
In this paper, we aim to bridge this gap and present a systematic study on the impact of multimodal fusion architecture design on 3D-AD. 
This work considers the multimodal fusion architecture design at the intra-module fusion level, i.e., independent modality-specific modules, involving early, middle or late multimodal features with specific fusion operations, and also at the inter-module fusion level, i.e., the strategies to fuse those modules. 
In both cases, we first derive insights through theoretically and experimentally exploring how architectural designs influence 3D-AD. 
Then, we extend SOTA neural architecture search (NAS) paradigm and propose 3D-ADNAS to simultaneously search across multimodal fusion strategies and modality-specific modules for the first time.
Extensive experiments show that 3D-ADNAS obtains consistent improvements in 3D-AD across various model capacities in terms of accuracy, frame rate, and memory usage,  and it exhibits great potential in dealing with few-shot 3D-AD tasks.
\end{abstract}

%

\section{Introduction\label{sec:intro}}
Industrial anomaly detection is expected to accurately find out the difference between normal samples and anomalies like human inspectors. To achieve this, an emerging way is to exploit both image color (RGB) and depth information (rather than only RGB) for quality inspection, termed as 3D anomaly detection (3D-AD) \cite{qin2023image,zavrtanik2024cheating}. It works well since 3D depth information has shown an essential role in improving industrial detection accuracy \cite{cao2024complementary}. The typical way of existing 3D-AD methods is to train a reconstruction-based model via restoring synthetic abnormal samples to normal ones, and then utilize reconstruction features for defect identification of discriminator \cite{zhou2024r3d}. One key challenge of this way is how to realize correct fusion of the two modalities (i.e., RGB images and 3D point cloud) through multimodal fusion network such that the integration of depth information will not interfere with color information \cite{gu2024rethinking}.  

Regarding to the multimodal fusion network, it is natural that consistently evolved architectures outperform the original ones. In this sense, we have to figure out a question that whether the existing multimodal fusion architectures are ideal for 3D-AD. As previewed in Figure 1, we can observe that the architectural designs (in terms of modality-specific modules and combination of these modules) have a significant impact on 3D-AD’s accuracy. In fact, this observation can be validated via theoretical analyses (as provided in \textbf{Revisiting 3D-AD Fusion Architecture Section}). This motivates us to design multimodal fusion architectures/topologies tailored for 3D-AD.


\begin{figure}[t!]
\begin{minipage}{0.52\linewidth}
		\centerline{\includegraphics[width=\textwidth]{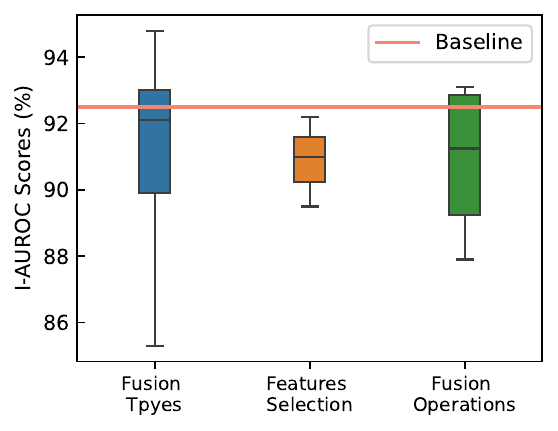}}
	\end{minipage}
	\begin{minipage}{0.46\linewidth}
		\centerline{\includegraphics[width=\textwidth]{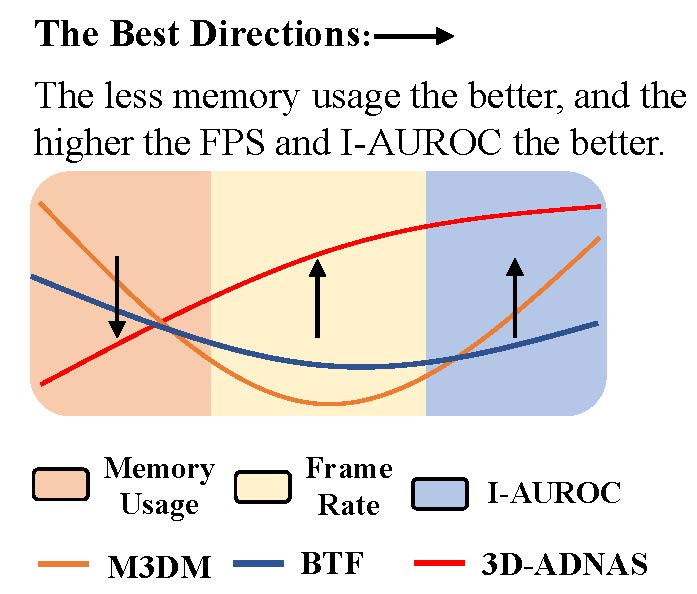}}
	\end{minipage}
  \label{Fig_11111}
  \caption{\textbf{(\textit{Left}) The impact of multimodal fusion architecture design} on  3D-AD performance. This shows the distribution of 3D-AD performance with variations at the intra- and inter-module fusion levels. \textbf{(\textit{Right}) 3D-ADNAS vs. SOTA methods} in terms of accuracy, FPS, and memory usage.}
\end{figure}

The primary goal of this work is to bridge the gap between multimodal fusion architectures and 3D-AD via (\romannumeral1) \textit{systematically studying the impact of architectural components on 3D-AD}, (\romannumeral2) \textit{recognizing crucial design schemes that can enhance 3D-AD}, and (\romannumeral3) \textit{proposing a simple yet effective 3D-ADNAS with a novel two-level search space tailored for 3D-AD}. We adopt empirical and theoretical approaches and conduct extensive experiments to realize this goal.


Given that EasyNet \cite{chen2023easynet} servers as foundation framework for 3D-AD, we initiate this work from it to revisit multimodal fusion network and formulate the design of target multimodal fusion network around a two-level multimodal fusion problem, i.e., intra-module fusion and inter-module fusion. Then we systematically assess the two main aspects of architecture design, intra-module fusion structure and inter-module fusion structure. Regarding to the former, we center around the investigation of modality-specific modules. Each module involves highly-correlated modality features, e.g., data-level or early features \cite{xu2021mufasa}; middle-level or hybrid features \cite{bergmann2023anomaly}; and classifier-level or late features \cite{wang2024m3dm}. Typical fusion operations include addition (weighted summation) and concatenation \cite{li2024towards,zhang2023multi}. Regarding to the latter, we seek to optimize fusion strategies, i.e., searching for optimal combination of those modules. To avoid randomness of the empirical observations, we repeat each ablation test multiple times with different random seeds. Then, we can obtain the following new observations:

\ding{182}  The single use of middle feature fusion is more favorable than early or late feature fusion, while the single use of late features degrades 3D-AD performance (\textbf{intra-module impact}).
 But, when combined with other fusion strategies, the late feature fusion can consistently improve performance across most 3D-AD tasks (\textbf{inter-module impact}). (Fig. \ref{Fig_2}-a)


\ding{183} Selecting the first two layers of middle-level features for fusion is, in general, more beneficial with 3D-AD training, as opposed to selecting all middle features used in standard 3D-AD (\textbf{intra-module impact}). (Fig. \ref{Fig_2}-b)

\ding{184}  Employing multiple fusion operations in each fusion module (i.e., early, middle, or late feature fusion module) is more effective than only one operation. Particularly, the guided attention operation and weighted summation operation are good at improving 3D-AD (\textbf{intra-module impact}). (Fig. \ref{Fig_2}-c and Table 1)


\ding{185} The multimodal fusion performance is highly dependent on the combination of early, middle and late features with relevant fusion operations. Especially, the automatic choice of the final fusion strategy according to the characteristics of features and operations is the core factor to realize 3D-AD (\textbf{inter-module impact}). (Fig. \ref{Fig_2}-a and Table 4)



\ding{186}
 In summary, the multimodal fusion architectures in terms of both intra- and inter-module aspects contribute enormously to  3D-AD.


 Based on these insights above, we can recognize key architectural designs that can boost the multimodal fusion of 3D-AD. Then, another question is arisen: \textbf{How to seek an efficient and 3D-AD-friendly multimodal fusion architecture?} A natural way is the Neural Architecture Search (NAS) technique \cite{liu2018darts}, which can perform an automated search over the architecture for 3D-AD. In this paper, we extend NAS paradigm for 3D-AD and boost multimodal fusion from an architectural perspective. Note that, the performance of NAS is sensitive to the design of multimodal fusion search space \cite{perez2019mfas,yin2022bm}. 
 Based on the investigation about the impact of architectural designs on 3D-AD’s performance, this work suggests a specialized two-level search space for 3D-AD, within which each architecture candidate can be concisely represented, facilitating an efficient search process.

The contributions of this work are as follows:
\begin{itemize}
    \item We thoroughly analyze the impact of multimodal fusion architectures on 3D-AD from both empirical and theoretical views, and recognize the crucial intra- and inter-module fusion designs that can boost 3D-AD performance.

    \item To the best of our knowledge, this work is the first attempt to utilize the NAS technique to shape a 3D-AD-friendly multimodal fusion architecture. To realize this, we design a 3D-ADNAS method to seek a promising fusion architecture across intra- and inter-module fusion strategies.

  \item  Extensive experiments validate the effectiveness of our design in improving anomaly detection of multimodal fusion network in terms of accuracy, speed and efficiency. Furthermore, our proposed method shows great potential in dealing with few-shot 3D-AD tasks.
\end{itemize}
\begin{figure*}[ht]
  \centering
  \setlength{\abovecaptionskip}{0.1cm}
      \includegraphics[scale=0.51]{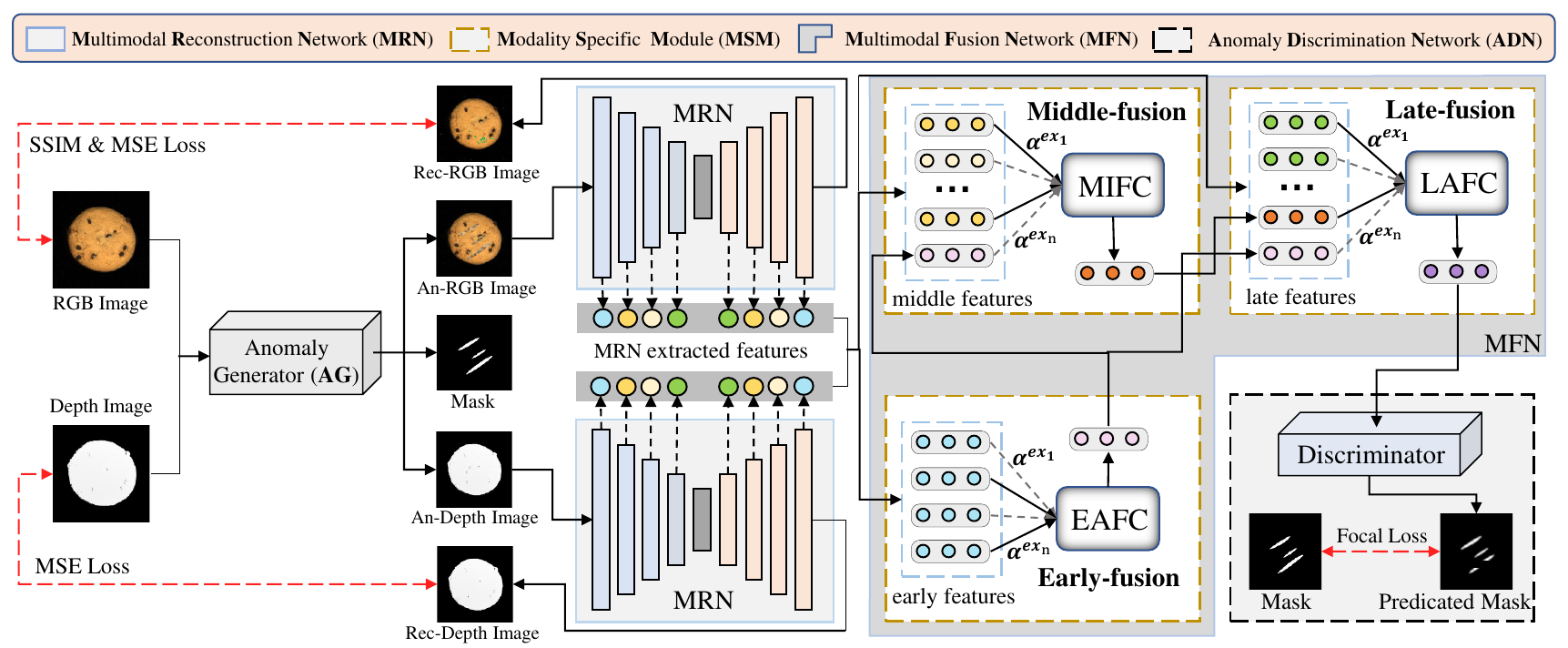}
  \caption{The overall framework of 3D-ADNAS, where the MFN architecture design is the core of this work, which is specified as a two-level search space: at the inter-module fusion level, the early fusion cell (EAFC), middle fusion cell (MIFC), and late fusion cell (LAFC) are configured to determine optimal combination of involved features and operations; at intra-module fusion level, it aims to seek best fusion strategy to combine those modules (MSMs).}
  \label{fig:3}
\end{figure*}

\section{Preliminaries}
In this section, we present the experimental setup for evaluating multimodal fusion architecture towards 3D-AD.
\vspace{0.05cm}

\textbf{Architecture Skeleton.}\quad Fig. \ref{fig:3} shows the skeleton of 3D-ADNAS, which consists of an Anomaly Generator (AG), a Multimodal Reconstruction Network (MRN), a Multimodal Fusion Network (MFN), and an Anomaly Discrimination Network (ADN). AG aims to simulate abnormal image generation according to the mask. MRN restores abnormal images (from AG) to normal ones and then extracts multi-scale features, which can be divided as early features, middle features and late features \cite{xu2021mufasa}. MFN consists of three types of independent modality-specific modules (MSMs), and it aims to fuse these multimodal features from MRN in a two-level manner. Then, ADN detects the products whether defective. Particularly, the two-level multimodal feature fusion of MFN is the target component of our design: (\romannumeral1) at \textbf{the intra-module fusion level}, we aim to optimize the inner structure of the three MSMs (i.e., early MSM, middle MSM and late MSM);  (\romannumeral2) at \textbf{the inter-module fusion level}, we optimize fusion strategies to combine those MSMs.
\vspace{0.05cm}

\textbf{MSM Structure.}\quad Fig. \ref{cell} shows the basic structure of an MSM, which consists of a specific candidate feature pool $\mathbb{F}$ (i.e, with early features, middle features or late features) and a fusion cell (i.e., EFAC, MIFC, or LAFC).  Each cell is regarded as a directed acyclic graph containing two input nodes, $\mathbb{K}$ intermediate nodes and an output node. It has three trainable architectural parameters that need to be optimized: $\alpha^{ex}$ denotes the weights of candidate features in the pool, $\alpha^{in}$ represents the weights of connections between input nodes and intermediate nodes, and $\beta^{op}$ denotes the weights of candidate operations (e.g., addition, concatenation, GLU, and guided attention) in the cell.

\textbf{Evaluation Metrics.}\quad We consider several popular metrics \cite{dai2024generating,sui2024cross} to evaluate the 3D-AD performance, including I-AUROC (image-level area under the receiver operating characteristic curve), P-AUROC (pixel-wise area under the receiver operating characteristic curve), and AUPRO (area under the per-region overlap curve).
\vspace{0.05cm}

\textbf{Implementation Details.}\quad For the tests about impact of intra/inter-module fusion architectures, we train the overall model of 3D-ADNAS by 600 epochs. For the tests about evaluation of searched architectures, we train MFN model by 80 epochs and follow the setting of DARTS to obtain best multimodal fusion model, and then train the overall model of 3D-ADNAS with obtained MFN by 600 epochs. 

\section{Revisiting 3D-AD Fusion Architecture} 
Then, we conduct experimentation and theoretical analysis to verify the impact of fusion architectures on 3D-AD.

\subsection{Impact of Inter-module Fusion}
The three MSMs all show effectiveness in multimodal learning \cite{tu2024self,wang2024incremental}. However, the impact of these fusion modules on 3D-AD remains unexplored. To bridge this gap, we conduct tests to scrutinize optimal combination of these MSMs on 3D-AD. For tests, we use EasyNet as the backbone for multimodal fusion of 3D-ADNAS. As shown in Fig. \ref{Fig_2}-a, we observe that these inter-module fusion strategies (i.e., the strategies about how to combine early, middle and late MSMs) indeed have significant impact on the performance of 3D-AD. For instance, when combined with middle MSM, the late MSM obtains obvious improvement of 3D-AD performance across most test tasks, but it shows harmfulness to 3D-AD if only using it itself. In most test cases, once the middle module is involved, the fusion module combination tends to have positive effect on 3D-AD, leading to significant performance improvement (see Table 4). However, this observation does not necessarily mean that the middle MSM is definitely superior to the others. So, we need to investigate the role of the intra-module fusion on 3D-AD.

\subsection{Impact of Intra-module Fusion} For each MSM, the fusion of its involved multimodal features (e.g., early, middle, or late features) with fusion operations (e.g., addition or concatenation \cite{gu2024anomalygpt}) is the key to realize intra-module fusion of 3D-AD. Thus, following the settings of above tests, we perform a series of experiments by selecting different multimodal features and fusion operations within a specific MSM to explore their impact on 3D-AD. Here, the middle MSM is used for test, which involves different stages of middle-level features. As shown in Fig. \ref{Fig_2}-b, we find that selecting partial layers of middle features is more effective for 3D-AD than selection all the features. From Fig. \ref{Fig_2}-c, we can see that the selection of fusion operations has important effect on 3D-AD, and employing multiple fusion operations is more beneficial with the fusion performance than only a single operation. Among these operations, the guided attention operation and weighted summation operation \cite{lei2023pyramidflow} show best effectiveness in 3D-AD tasks. These results validate the actual impact of intra-module fusion design on 3D-AD.

\begin{figure}[t]
\centering
      \includegraphics[scale=0.48]{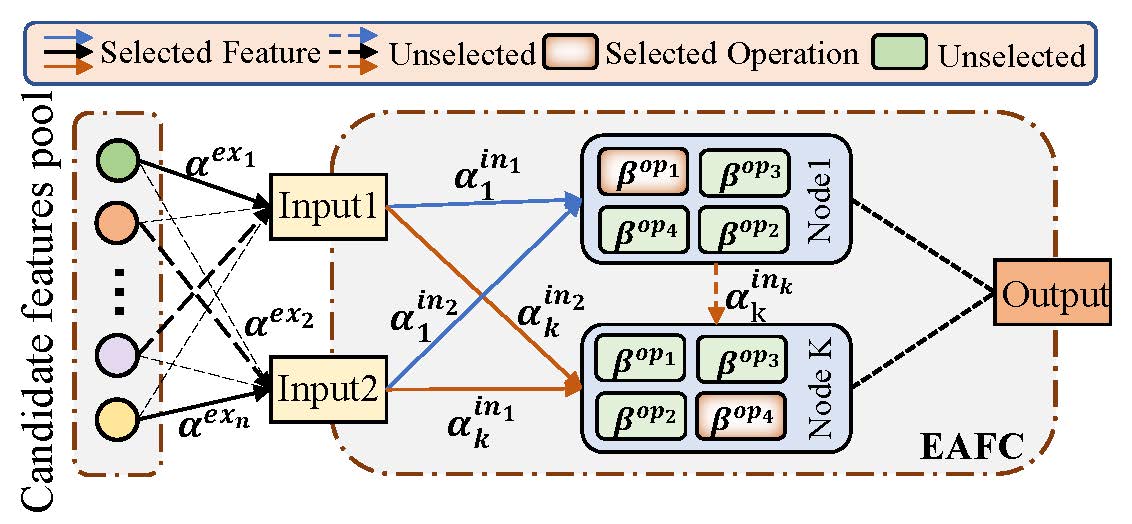}
  \caption{The inner structure of an MSM, where the early MSM is used as example with $\mathbb{K}=2$. Note that the three types of MSMs share a similar structure.}
 \label{cell}
\end{figure}
\subsection{Theoretical Analysis}
Further, we theoretically analyze the impact of our target intra- and inter-module fusion design on 3D-AD using the Dempster-Shafer’s evidence theory (DST) \cite{liu2017weighted,DBLP:conf/iclr/HanZFZ21}. In the following, we first consider the impact of inter-module fusion design, where
the DST's concepts including
belief mass $b$ and uncertainty $u$ are utilized to evaluate the trustworthiness of target fusion model’s outputs. 

For an $N$-class classification task of MSM, given the opinion of original MSM (e.g., the late MSM $l$): $L=\{\{b_l^n\}_{n=1}^N,u_l\}$, we aim to theoretically analyze whether the fusion of additional opinion (e.g., the middle MSM $m$) $M=\{\{b_m^n\}_{n=1}^N,u_m\}$ will influence the model's classification accuracy.
Then, following the combination rule of DST, we fuse $L$ into $M$ and form a new opinion $F=\{\{b_f^n\}_{n=1}^N,u_f\}$. Here, $b_{f}^n$ and $u_f$ are new belief mass and uncertainty, respectively, which are given by
\begin{equation}
\begin{gathered}
b_{f}^n=(b_l^{n} b_m^{n} +b_l^{n} u_m + b_m^n u_l) / {(1-z)},\\
u_{f}=(u_l u_m) / {(1-z)},
\end{gathered}
\end{equation}
\noindent where $z=\sum_{i\neq j} b_l^i b_m^j$ ($i,j \in [1,2, ..., N]$) is the measure of the conflict quantity between two belief mass sets of $L$ and $M$, and $\frac{1}{1-z}$ is used as the normalization factor.

Then, we can give the following propositions (more detailed proofs are provided in the Supplementary Material\footnote{Please refer to the following for additional material: https://github.com/longkaifang/3D-ADNAS.}).

\vspace{0.15cm}
\noindent\fbox{%
	\parbox{0.48\textwidth}{\textit{\textbf{Proposition}} \textbf{1.} Under the conditions $b_m^g \geq b_l^{max}$, where $g \in N$ is the index of the ground-truth label, and $b_l^{max}$ is the largest in $\{b_l^n\}_{n=1}^N$, fusing another opinion $M$ makes the new opinion $F$ satisfy $b_f^g \geq b_l^g$.
}}
\\

\textit{\textbf{Proof.}}
\begin{equation}
\begin{split}
b_{f}^{g}&= \frac{b_l^{g} b_m^{g} +b_l^{g} u_m + b_m^g u_l}{\sum_{n=1}^N b_m^n b_l^n +u_m +u_l -u_m u_l}\\
& \geq \frac{b_l^{g} b_m^{g} +b_l^{g} u_m + b_l^{max} u_l}{\sum_{n=1}^N b_m^n b_l^{max} +u_m +u_l -u_m u_l}\\
& \geq b_e^g \frac{ (b_l^{max} + u_m + u_l)}{b_l^{max} + u_m + u_l} \geq b_l^g.\\
\end{split}
\end{equation}

\noindent\fbox{
	\parbox{0.48\textwidth}{\textit{\textbf{Proposition}} \textbf{2.} When $u_m$ is large, $b_l^g$ $-$ $b_f^g$ will be limited, and it will have a negative correlation with $u_m$. As a special case, when $u_m$ is large enough (i.e., $u_m$ = 1), fusing another opinion will not reduce the performance (i.e., $b_f^g = b_l^g$ ).
}}\\


\textit{\textbf{Proof.}}
\begin{equation}
\begin{split}
b_l^g - b_f^g&=b_l^g-\frac{b_l^{g} b_m^{g} +b_l^{g} u_m + b_m^g u_l}{\sum_{n=1}^N b_m^n b_l^n +u_m +u_l -u_m u_l}\\
&\leq b_l^g-\frac{b_l^g u_m}{b_m^{max} \cdot 1 +u_m +u_l -u_m u_l}\\
&\leq b_l^g-\frac{b_l^g u_m}{1 +u_l -u_m u_l} =b_l^g \frac{1 +u_l}{\frac{1}{(1-u_m)} +u_l},
\end{split}
\end{equation}

To this end, we have the following conclusions: (\romannumeral1) According to \textbf{\textit{Proposition} 1}, fusing an additional opinion (e.g., $M$) into the original opinion (e.g., $L$) has great potential to boost the model's accuracy. (\romannumeral2) According to \textbf{\textit{Proposition} 2}, the above fusion may lead to accuracy deterioration, but even this is limited under mild conditions.

For the impact of intra-module design, we can utilize the same analysis approach as mentioned above, and obtain the same conclusions as the case of the above inter-module design, which are provided in the Supplementary Material. 

\begin{figure}[t]
\begin{minipage}{0.52\linewidth}
		\centerline{\includegraphics[width=\textwidth]{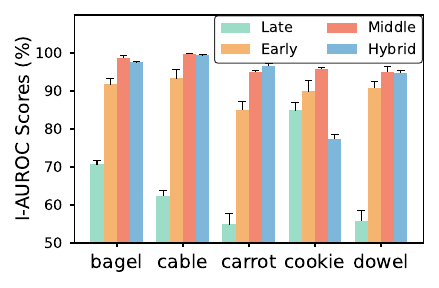}}
	\end{minipage}
	\begin{minipage}{0.46\linewidth}
		\centerline{\includegraphics[width=\textwidth]{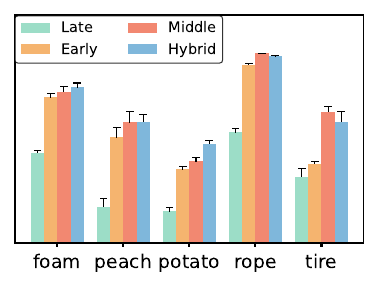}}
	\end{minipage}
 \centerline{(a) Impact of fusion types.}
  \begin{minipage}{0.52\linewidth}
		\centerline{\includegraphics[width=\textwidth]{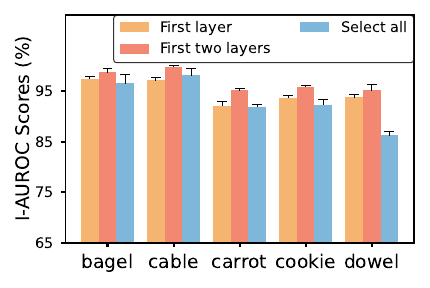}}
	\end{minipage}
	\begin{minipage}{0.46\linewidth}
		\centerline{\includegraphics[width=\textwidth]{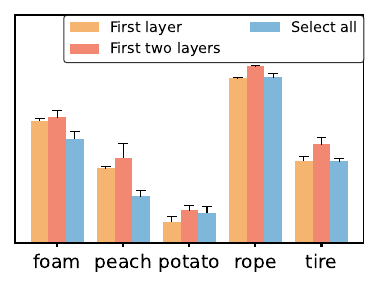}}
	\end{minipage}
 \centerline{(b) Impact of features selection.}
	\begin{minipage}{0.52\linewidth}
		\centerline{\includegraphics[width=\textwidth]{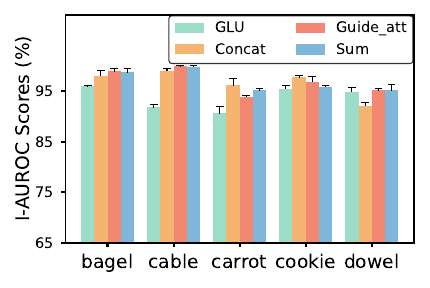}}
	\end{minipage}
	\begin{minipage}{0.46\linewidth}
		\centerline{\includegraphics[width=\textwidth]{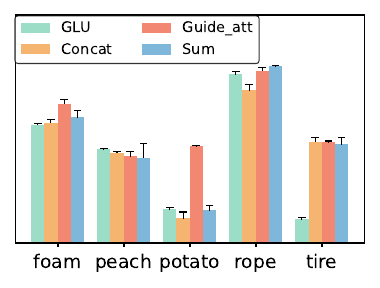}}
	\end{minipage}
 \centerline{(c) Impact of fusion operations.}
	\caption{The impact of multimodal fusion architecture design on 3D-AD performance. Zoom in for details.}
  \label{Fig_2}
\end{figure}
\begin{table*}[t!]
\centering
\footnotesize
\setlength{\tabcolsep}{0.55em}
\renewcommand{\arraystretch}{1}
\adjustbox{width=0.9\linewidth}{
   \begin{tabular}{c|l|c|c ccc ccc ccc |c}
       \toprule
                 \multirow{1}{*}{}
                & \multirow{2}{*}{Method}
                & \multirow{2}{*}{Year}
                &\multirow{1}{*}{Candy}
                &\multirow{1}{*}{Chocolate} 
                &\multirow{1}{*}{Chocolate} 
                &\multirow{2}{*}{Confetto} 
                &\multirow{1}{*}{Gummy} 
                &\multirow{1}{*}{Hazelnut} 
                &\multirow{1}{*}{Licorice} 
                &\multirow{2}{*}{Lollipop} 
                &\multirow{1}{*}{Marsh-} 
                &\multirow{1}{*}{Peppermint}
                 &\multirow{2}{*}{Mean} \\
                && &Cane  &Cookie  &Praline  &  &Bear  &Truffle  &Sandwish  &   &mallow  &Candy  \\
\midrule[0.5pt]  
\multirow{3}{*}{\rotatebox{90}{RGB}}            
                     &EasyNet &MM23 &\color{blue}{72.3} &92.5 & 84.9&\color{blue}{96.6} &70.5 &\color{blue}{81.5} &80.6 &\color{blue}{85.1} & 97.5 & \color{blue}{96.0} &85.8\\
                     &M3DM &CVPR23 &64.8 & \color{blue}{94.9} & \color{red}{94.1} &\color{red}{100.0} &\color{red}{87.8} &63.2 &\color{red}{93.3} &81.1 &\color{red}{99.8} &\color{red}{100.0} &\color{blue}{87.9}\\
              & \cellcolor{gray!20}3D-ADNAS &\cellcolor{gray!20}- & \cellcolor{gray!20}\color{red}{79.8} & \cellcolor{gray!20}\color{red}{99.8} & \cellcolor{gray!20}\color{blue}{92.6} &\cellcolor{gray!20}\color{red}{100.0}& \cellcolor{gray!20}\color{blue}{83.5}& \cellcolor{gray!20}\color{red}{86.2}& \cellcolor{gray!20}\color{blue}{91.2}& \cellcolor{gray!20}\color{red}{96.0}& \cellcolor{gray!20}\color{blue}{97.8}& \cellcolor{gray!20}\color{red}{100.0}& \cellcolor{gray!20}\color{red}{92.7}\\
\midrule[0.5pt]
\multirow{3}{*}{\rotatebox{90}{3D}}          
                     &EasyNet  &MM23 &\color{red}{62.9} & \color{red}{71.6} &76.8 & 73.1 &\color{blue}{66.0} & \color{red}{71.0} & 71.2 & 71.1 & 68.8 & 73.1 & 70.6\\
                     &M3DM &CVPR23 &48.2 & 58.9 & \color{blue}{80.5} &\color{red}{84.5} &\color{red}{78.0} &53.8 & \color{blue}{76.6}&\color{red}{82.7} &\color{blue}{80.0} & \color{blue}{82.2} &\color{blue}{72.5}\\            
              & \cellcolor{gray!20}3D-ADNAS &\cellcolor{gray!20}- & \cellcolor{gray!20}\color{blue}{53.3}& \cellcolor{gray!20}\color{blue}{65.6}& \cellcolor{gray!20}\color{red}{85.9}& \cellcolor{gray!20}\color{blue}{79.5} &\cellcolor{gray!20}\color{red}{78.0}& \cellcolor{gray!20}\color{blue}{62.9}& \cellcolor{gray!20}\color{red}{84.3}& \cellcolor{gray!20}\color{blue}{78.1}& \cellcolor{gray!20}\color{red}{82.4}& \cellcolor{gray!20}\color{red}{87.8} & \cellcolor{gray!20}\color{red}{75.8}\\
\midrule[0.5pt]
\multirow{5}{*}{\rotatebox{90}{RGB + 3D}}
                      &EasyNet  &MM23 &\color{blue}{73.7} & 93.4 &86.6 &\color{blue}{96.6} &71.7 & \color{blue}{82.2} & 84.7 &\color{blue}{86.3} & 97.7 & 96.0 & 86.9\\
                     &AST &WACV23 &57.4 & 74.7 &74.7 &88.9 &59.6 & 61.7 &81.6 &84.1 &98.7& \color{blue}{98.7} &78.0\\
                     &M3DM &CVPR23 &62.4 & \color{blue}{95.8} &\color{blue}{ 95.8} &\color{red}{100.0} &\color{blue}{88.6} & 78.5 &\color{red}{94.9} & 83.6 &\color{red}{100.0} & \color{red}{100.0} & \color{blue}{89.7}\\
                     &CFM &CVPR24 &68.0 & 93.1 & 95.2 & 88.0 &\color{red}{86.5}  &78.2 &91.7 &84.0 &\color{blue}{99.8} &96.2&88.1\\
              & \cellcolor{gray!20}3D-ADNAS &\cellcolor{gray!20}- & \cellcolor{gray!20}\color{red}{89.6}& \cellcolor{gray!20}\color{red}{100.0}& \cellcolor{gray!20}\color{red}{97.0}& \cellcolor{gray!20}\color{red}{100.0}& \cellcolor{gray!20}82.7& \cellcolor{gray!20}\color{red}{88.2}& \cellcolor{gray!20}\color{blue}{93.1}& \cellcolor{gray!20}\color{red}{95.0}& \cellcolor{gray!20}\color{red}{100.0}& \cellcolor{gray!20}\color{red}{100.0}& \cellcolor{gray!20}\color{red}{94.6}\\
             \bottomrule
       \end{tabular}
   }
\caption{I-AUROC scores on  Eyecandies dataset. The red indicates the best results and the blue indicates the second best.}
\end{table*}
\begin{table*}[t!]
\centering
\footnotesize
\setlength{\tabcolsep}{0.55em}
\renewcommand{\arraystretch}{1}
\adjustbox{width=0.9\linewidth}{
   \begin{tabular}{c|l|c|c ccc ccc ccc |c}
       \toprule
                 \multirow{1}{*}{}
                & \multirow{2}{*}{Method}
                & \multirow{2}{*}{Year}
                &\multirow{1}{*}{Candy}
                &\multirow{1}{*}{Chocolate} 
                &\multirow{1}{*}{Chocolate} 
                &\multirow{2}{*}{Confetto} 
                &\multirow{1}{*}{Gummy} 
                &\multirow{1}{*}{Hazelnut} 
                &\multirow{1}{*}{Licorice} 
                &\multirow{2}{*}{Lollipop} 
                &\multirow{1}{*}{Marsh-} 
                &\multirow{1}{*}{Peppermint}
                 &\multirow{2}{*}{Mean} \\
                && &Cane  &Cookie  &Praline  &  &Bear  &Truffle  &Sandwish  &   &mallow  &Candy  \\
\midrule[0.5pt]  
\multirow{2}{*}{\rotatebox{90}{RGB} }    
                     &M3DM& CVPR23 &\color{blue}{86.7} &\color{red}{90.4} & \color{blue}{80.5}&\color{red}{98.2} &\color{red}{87.1} &\color{blue}{66.2} &\color{blue}{88.2} &\color{red}{89.5} & \color{red}{97.0} & \color{blue}{96.2} &\color{blue}{88.0}\\
              & \cellcolor{gray!20}3D-ADNAS & \cellcolor{gray!20}-& \cellcolor{gray!20}\color{red}{89.7} & \cellcolor{gray!20}\color{blue}{87.6} & \cellcolor{gray!20}\color{red}{84.7} &\cellcolor{gray!20}\color{blue}{97.1}& \cellcolor{gray!20}\color{blue}{79.6}& \cellcolor{gray!20}\color{red}{75.4}& \cellcolor{gray!20}\color{red}{90.1}& \cellcolor{gray!20}\color{blue}{85.3}& \cellcolor{gray!20}\color{blue}{95.1}& \cellcolor{gray!20}\color{red}{96.9}
              & \cellcolor{gray!20}\color{red}{88.2}\\
\midrule[0.5pt]
\multirow{2}{*}{\rotatebox{90}{{3D}} }         
                     &M3DM& CVPR23  &\color{red}{91.1} & \color{blue}{64.5} &\color{red}{58.1} & \color{red}{74.8} &\color{red}{74.8} & \color{red}{48.4} & \color{red}{60.8} & \color{red}{90.4} &\color{blue}{64.6} &\color{blue}{75.0}& \color{red}{70.2}\\
              & \cellcolor{gray!20}3D-ADNAS &\cellcolor{gray!20}-& \cellcolor{gray!20}\color{blue}{88.2}& \cellcolor{gray!20}\color{red}{66.7}& \cellcolor{gray!20}\color{blue}{54.2}& \cellcolor{gray!20}\color{blue}{55.8 }&\cellcolor{gray!20}\color{blue}{63.9}& \cellcolor{gray!20}\color{blue}{45.4}& \cellcolor{gray!20}\color{blue}{53.8}& \cellcolor{gray!20}\color{blue}{74.6}& \cellcolor{gray!20}\color{red}{70.1}& \cellcolor{gray!20}\color{red}{83.6} & \cellcolor{gray!20}\color{blue}{65.7}\\
\midrule[0.5pt]
\multirow{4}{*}{\rotatebox{90}{{RGB+3D}}} 
                      &AST& WACV23   &51.4 & 83.5 &71.4 &90.5 &58.7 & 59.0 & 73.6 &76.9 & 91.8 & 87.8 & 74.4\\
                      &M3DM& CVPR23   &90.6 & \color{red}{92.3} &80.3 &\color{red}{98.3} &85.5 & 68.8 & \color{blue}{88.0} &90.6 & \color{red}{96.6} & \color{blue}{95.5} & 88.2\\
                      &CFM& CVPR24  &\color{blue}{94.2} & \color{blue}{90.2} &\color{red}{83.1} &\color{blue}{96.5} &\color{red}{87.5} & \color{red}{76.2} & 79.1 &\color{red}{91.3} & 93.9 & 94.9 & \color{blue}{88.7}\\
              & \cellcolor{gray!20}3D-ADNAS  &\cellcolor{gray!20}-& \cellcolor{gray!20}\color{red}{94.5}& \cellcolor{gray!20}89.1& \cellcolor{gray!20}\color{blue}{82.7}& \cellcolor{gray!20}95.8& \cellcolor{gray!20}\color{blue}{85.7}& \cellcolor{gray!20}\color{blue}{74.8}& \cellcolor{gray!20}\color{red}{91.1}& \cellcolor{gray!20}\color{blue}{90.7}& \cellcolor{gray!20}\color{blue}{96.4}& \cellcolor{gray!20}\color{red}{97.2}& \cellcolor{gray!20}\color{red}{89.8}\\
             \bottomrule
       \end{tabular}
   }
\caption{AUPRO scores on  Eyecandies dataset. The red indicates the best results and the blue indicates the second best.}
\end{table*}
\section{3D-ADNAS Method}
To seek optimal architectural designs towards 3D-AD, we propose a simple yet powerful NAS method (3D-ADNAS) with a two-level search space and a simple search strategy grounded in gradient-based algorithms.

\subsection{Search Space towards 3D-ADNAS}
The above experimentation and theoretical analysis show the important role of the intra-module and inter-module fusion architectures in the 3D-AD skeleton. Therefore, to achieve a better 3D-AD performance, further exploration of the above two-level architectural design is necessary, i.e., the two-level design of 3D-AD-friendly search space.
\vspace{0.08cm}

\noindent\textbf{Intra-module fusion level.}\quad The impact of various components, i.e., module-specific features (e.g., early, middle or late features), and  fusion operations of the cell in
each MSM, has not been thoroughly examined in prior experiments. Then, the optimal designs of the combination of these components, are necessary to search for, which needs to consider the following aspects. 

\textbf{What features are selected as cell inputs?} 
As shown in Fig. \ref{cell}, for each MSM, we need to select two features from the candidate feature pool ($\mathbb{F}$) as the inputs of the cell according to architectural parameters $\alpha^{ex}$. That is, it is required to optimize $\alpha^{ex}$ and select the features with largest $\alpha^{ex}$ values as optimal solutions.  To solve this problem, we utilize the continuous relaxation strategy of DARTS to convert the discrete feature selection problem into the continuous search problem, and then use gradient method to solve it. To achieve this, we reformulate the cell inputs through weighted summation of all candidate multimodal features in $\mathbb{F}$ as:
\begin{equation}
\Tilde{X}_i=\sum_{f_s\in \mathbb{F}} \frac{exp(\alpha^{ex_i}_i)}{\sum_{j=1}^\mathbb{F} exp(\alpha_i^{{ex_j}})} \cdot {f_s},
\end{equation}
where $\Tilde X_i$ denotes the input features of each cell, $i \in [1,2]$, $f_s$ denotes the feature in $\mathbb{F}$. Accordingly, the two features serving as inputs of each cell's can be determined by:
\begin{equation}
(f_s^j,f_s^h)=\mathop{\arg\max}\limits_{i\in [1,\mathbb{F}]} \, (\alpha_1^{ex_i},\alpha_2^{ex_i}),
\end{equation}
where $(f_s^j,f_s^h)$ denotes the choice of the $j$-th and $h$-th features from $\mathbb{F}$ as cell inputs.

Similarly, following the above approach, we can also obtain the optimal connections (identified by $\alpha^{in}$) between input nodes and intermediate nodes in each cell. More details are provided in the Supplementary Material. 

\textbf{What operations are selected for features fusion?}  
To achieve best fusion of the above input features, we need to select appropriate fusion operations from the candidate operation fool $\mathbb{O}$ (including addition, concatenation, GLU, and guided attention) for each intermediate node according to architectural parameters $\beta^{op}$. Similar to the case of feature selection, we relax the operations in $\mathbb{O}$ to obtain the output of $k$-th intermediate node during the search, as follows:
\begin{equation}
\Tilde{op} (t^y,t^z)=\sum_{op\in \mathbb{O}} \frac{exp(\beta^{op})}{\sum_{op^{\prime}\in \mathbb{O}} exp(\beta^{{op^\prime}})} op (t^y,t^z),
\end{equation}
where $op$ denotes the primitive operation in $\mathbb{O}$, $t^y$ and $t^z$ represent the input features of $k$-th intermediate node. Then, after gradient-based search, the best fusion operations for each intermediate node can be obtained by discretizing $\beta^{op}$ as follows:
\begin{equation}
op(\cdot)=\mathop{\arg\max} \,\beta^{op}.
\end{equation}
\noindent\textbf{Inter-module fusion level.}\quad After the optimization of the MSMs at the intra-module fusion level, a natural question is arisen: \textbf{how to combine those  MSMs to achieve best final fusion performance?} For this issue, we formulate a simple yet effective inter-module fusion-level search space, where the output of early MSM is considered as a candidate input feature in the feature pool of the other two MSMs, the middle MSM output is considered as a candidate input feature of late MSM, and it is assigned with a weight $\alpha^{ex}$, which can be optimized by gradient method. Finally, the output features of late MSM together with reconstructed image feature are fed into ADN for anomaly detection. In this way, we can search for optimal fusion strategy to integrate these MSNs in an efficient manner.


\begin{table*}[t!]
\centering
\footnotesize
\setlength{\tabcolsep}{2.8mm} 
\renewcommand{\arraystretch}{1}
\adjustbox{width=0.9\linewidth}{
   \begin{tabular}{c|l|c|c ccc ccc ccc |c}
       \toprule
                 \multirow{1}{*}{}
                 & \multirow{2}{*}{Method}
                 & \multirow{2}{*}{Year}
                &\multirow{2}{*}{Bagel}
                &\multirow{2}{*}{Cable Gland} 
                &\multirow{2}{*}{Carrot} 
                &\multirow{2}{*}{Cookie} 
                &\multirow{2}{*}{Dowel} 
                &\multirow{2}{*}{Foam} 
                &\multirow{2}{*}{Peach} 
                &\multirow{2}{*}{Potato} 
                &\multirow{2}{*}{Rope} 
                &\multirow{2}{*}{Tire}
                 &\multirow{2}{*}{Mean} \\
                & &&  &  &  &  &  &  &  &   &  &  \\
\midrule[0.5pt]  
\multirow{3}{*}{\rotatebox{90}{RGB}}               
                     &EasyNet  &MM23 &\color{red}{98.2}&\color{red}{99.2} & \color{blue}{91.7}&\color{blue}{95.3} &91.9 &\color{blue}{92.3} &84.0&\color{blue}{78.5} & \color{red}{98.6} & 74.2 &\color{blue}{90.4}\\
                     &M3DM &CVPR23 &94.4 & 91.8 & 89.6 &74.9 &\color{red}{95.9} &76.7 &\color{red}{91.9} &64.8 &93.8 &\color{blue}{76.7} &85.0\\
              & \cellcolor{gray!20}3D-ADNAS &\cellcolor{gray!20}- & \cellcolor{gray!20}\color{blue}{98.1} & \cellcolor{gray!20}\color{blue}{98.8} & \cellcolor{gray!20}\color{red}{92.7}& \cellcolor{gray!20}\color{red}{95.6}& \cellcolor{gray!20}\color{blue}{94.2}& \cellcolor{gray!20}\color{red}{92.8}& \cellcolor{gray!20}\color{blue}{85.3}& \cellcolor{gray!20}\color{red}{79.1}& \cellcolor{gray!20}\color{blue}{97.7}& \cellcolor{gray!20}\color{red}{85.8} & \cellcolor{gray!20}\color{red}{92.0}\\
\midrule[0.5pt]
\multirow{3}{*}{\rotatebox{90}{3D}}                 
                     &EasyNet  &MM23 &73.5 & \color{blue}{67.8} & \color{blue}{74.7} & 86.4 &\color{blue}{71.9} & \color{blue}{71.6} & \color{blue}{71.3} & 72.5 & 88.5 & \color{blue}{68.7} & 74.7\\
                     &M3DM &CVPR23 &\color{red}{94.1} & 65.1 & \color{red}{96.5} &\color{red}{96.9} &\color{red}{90.5} & \color{red}{76.0} & \color{red}{88.0} &\color{red}{97.4} &\color{blue}{92.6} & \color{red}{76.5} &\color{red}{87.4}\\            
              & \cellcolor{gray!20}3D-ADNAS & \cellcolor{gray!20}- & \cellcolor{gray!20}\color{blue}{79.4} & \cellcolor{gray!20}\color{red}{85.7} & \cellcolor{gray!20}69.9& \cellcolor{gray!20}\color{blue}{94.6}& \cellcolor{gray!20}69.5& \cellcolor{gray!20}68.6& \cellcolor{gray!20}70.5& \cellcolor{gray!20}\color{blue}{87.3}& \cellcolor{gray!20}\color{red}{95.3}& \cellcolor{gray!20}66.7 & \cellcolor{gray!20}\color{blue}{78.8}\\
\midrule[0.5pt]
\multirow{6}{*}{\rotatebox{90}{RGB + 3D}}
                      &BTF &CVPR23 &93.8 & 76.5 & 97.2 & 88.8 &96.0 & 66.4 & 90.4 & 92.9 &98.2 & 72.6 & 87.3\\
                      &EasyNet  &MM23 &99.1 & \color{blue}{99.8} & 91.8 &96.8 & 94.5&\color{blue}{94.5}  & 90.5 &80.7 & \color{blue}{99.4} & 79.3 & 92.6\\
                     &AST &WACV23 &98.3 & 87.3 &\color{blue}{97.6} &97.1 &93.2 & 88.5 &\color{blue}{97.4} &\color{red}{98.1} &\color{red}{100.0}& 79.7 &93.7\\
                     &M3DM &CVPR23 &\color{blue}{99.4} & 90.9 & 97.2 &97.6 &96.0 & 94.2 &97.3 & 89.9 &97.2 & 85.0 & 94.5\\
                       &Shape\_Guided  &ICML23 &98.6 & 89.4 & \color{red}{98.3} & \color{red}{99.1} &\color{red}{97.6} & 85.7 & \color{red}{99.0} &\color{blue}{96.5} & 96.0 & \color{red}{86.9} &\color{blue}{94.7}\\
              & \cellcolor{gray!20}3D-ADNAS & \cellcolor{gray!20}- & \cellcolor{gray!20} \color{red}{99.7} & \cellcolor{gray!20}\color{red}{100.0} & \cellcolor{gray!20}97.1& \cellcolor{gray!20}\color{blue}{98.6}& \cellcolor{gray!20}\color{blue}{96.6}& \cellcolor{gray!20}\color{red}{94.8}& \cellcolor{gray!20}89.7& \cellcolor{gray!20}87.3& \cellcolor{gray!20}\color{red}{100.0}& \cellcolor{gray!20}\color{blue}{86.7} & \cellcolor{gray!20}\color{red}{95.1}\\
             \bottomrule
       \end{tabular}
   }
\caption{I-AUROC scores on  MVTec 3D-AD dataset. The red indicates the best results and the blue indicates the second best.}
\end{table*}
\begin{table*}[t!]
\centering
\tiny
\setlength{\tabcolsep}{2.2mm}
\renewcommand{\arraystretch}{1}
\adjustbox{width=0.9\linewidth}{
   \begin{tabular}{ccc|c ccc ccc ccc |c}
       \toprule
                 \multicolumn{3}{c|}{MSM fusion components}  
                &\multirow{2}{*}{Bagel}
                &\multirow{2}{*}{Cable Gland} 
                &\multirow{2}{*}{Carrot} 
                &\multirow{2}{*}{Cookie} 
                &\multirow{2}{*}{Dowel} 
                &\multirow{2}{*}{Foam} 
                &\multirow{2}{*}{Peach} 
                &\multirow{2}{*}{Potato} 
                &\multirow{2}{*}{Rope} 
                &\multirow{2}{*}{Tire}
                &\multirow{2}{*}{Mean}\\ 
                   Early& Middle& Late & & &  &  &  &  &  &  &   &  &  \\
\midrule[0.5pt]           
                     \Checkmark&\XSolidBrush&\XSolidBrush &89.6 &98.1 & 83.5 &87.1 &85.8 &89.7 &74.3 & \color{blue}{78.1} &96.5 &70.3 &85.3\\
                     \XSolidBrush&\Checkmark&\XSolidBrush &\color{blue}{99.5} &\color{blue}{99.9} & 94.5 &\color{blue}{95.6} &\color{blue}{96.2} &91.6 &85.7 & 74.0 &\color{red}{100.0} & 83.7 &92.1\\
                     \XSolidBrush&\XSolidBrush& \Checkmark&69.9 &60.9 & 54.5 &82.6 &53.3 &73.9 &57.0 & 67.9 &77.9 & 66.3 &66.4\\
                     \Checkmark&\Checkmark&\XSolidBrush &99.4 &\color{red}{100.0} & 92.9 &95.4 &93.9 &\color{blue}{93.3} &\color{blue}{84.7} & 83.8 &\color{blue}{99.8} & \color{red}{87.1} &\color{blue}{93.0}\\
                     \Checkmark &\XSolidBrush&\Checkmark &83.6 &79.5 &70.1 &85.9 &66.7 &77.5 &58.9 & 54.5 &76.3 & 57.8 &71.1\\
                     \XSolidBrush&\Checkmark &\Checkmark &97.4 &\color{red}{100.0} & \color{blue}{96.2} &78.5 &94.3 &89.8 &82.9 & 76.9 &98.8 & 84.6 &89.9\\
                     \Checkmark& \Checkmark& \Checkmark &\color{red}{99.7} &\color{red}{100.0} & \color{red}{97.1} &\color{red}{98.6} &\color{red}{96.6} &\color{red}{94.8} &\color{red}{89.7} & \color{red}{87.3}& \color{red}{100.0} &\color{blue}{86.7}  &\color{red}{95.1}\\

\bottomrule
\end{tabular}
   }
\caption{Ablation study on different fusion components. The red indicates the best results and the blue shows the second best.}
\end{table*}
\begin{table}[t]\normalsize
    \centering
    \renewcommand\arraystretch{1} 
    \setlength{\tabcolsep}{5.0mm}{
    \resizebox{\linewidth}{!}{
    	{\begin{tabular}{l|c|c|cc}
    		\toprule[1.0pt]
    	\multirow{2}{*}{Method}  &\multicolumn{3}{c}{MVTec 3D-AD} \\ 
    		&Frame Rate &Memory &I-AUROC \\ 
                \midrule[0.5pt]
            BTF &3.197 	&381.06  &86.5   \\
            AST &4.966 	&463.94  &93.7   \\
            M3DM &0.514 	&6526.12  &94.5 \\
            
            \rowcolor{gray!20} 3D-ADNAS (Ours) &\textbf{24.693}	&\textbf{269.33}	&\textbf{95.1}\\
            \bottomrule[1.0pt]
    \end{tabular}}}}
    \caption{Comparison results in terms of frame rate, memory usage and accuracy metrics.}\label{a_cost}
\end{table}
\begin{table}[t!]
    \centering
    \renewcommand\arraystretch{1}
    \setlength{\tabcolsep}{5.0mm}{
    \resizebox{\linewidth}{!}{
    	{\begin{tabular}{l|ccccc}
    		\toprule[1.0pt]
    	\multirow{2}{*}{Metrics} &\multicolumn{4}{c}{Eyecandies} \\ 
    		&5-shot &10-shot &50-shot &Full \\ 
                \midrule[0.5pt]
           I-AUROC & 77.5 &80.7 &86.8 &94.6\\
            P-AUROC & 87.5 &86.9 &91.2 &97.0\\
            AUPRO & 70.4 &76.7 &83.5 &89.8\\
            \bottomrule[1.0pt]
    \end{tabular}}}}
    \caption{Few-shot test results on Eyecandies.} \label{a_candidate}
\end{table}
\begin{table}[t]
    \centering
    \renewcommand\arraystretch{1}
    \setlength{\tabcolsep}{5.0mm}{
    \resizebox{\linewidth}{!}{
    	{\begin{tabular}{l|ccccc}
    		\toprule[1.0pt]
    	\multirow{2}{*}{Method} &\multicolumn{4}{c}{MVTec 3D-AD (I-AUROC)} \\ 
    		&5-shot &10-shot &50-shot &Full \\ 
                \midrule[0.5pt]
            BTF & 67.1 &69.5 &80.6 &86.5\\
            AST & 68.0 &68.9 &79.4 &93.7\\
            M3DM & 82.2 &84.5 &\textbf{90.7} &94.5\\
            \rowcolor{gray!20} 3D-ADNAS (Ours) 	&\textbf{82.6} &\textbf{84.8} &89.0
            &\textbf{95.1 }\\
            \bottomrule[1.0pt]
    \end{tabular}}}}
    \caption{Few-shot test results on MVTec 3D-AD.} \label{a_candidate1}
\end{table}
\subsection{Search  Strategy towards 3D-ADNAS}
Since our relaxed continuous search space is differentiable, we can employ the popular gradient-based method to alternately optimize the architectural parameters ($\alpha^{ex}$, $\alpha^{in}$ and $\beta^{op}$) and network weights ($w$) of 3D-ADNAS until the training of model converges.
When the search is completed, we derive the optimal multimodal fusion architecture according to the values of the learned architectural parameters. In this work, for simplicity, we adopt the basic gradient descent method used in DARTS. Note that any other more effective gradient-based search paradigms can be applied in our target scenario for composite benefit. The detailed algorithm of 3D-ADNAS is provided in the Supplementary Material.

\subsection{Performance Evaluation towards 3D-ADNAS} 
We employ the I-AUROC, P-AUROC, and AUPRO \cite{tien2023revisiting} metrics to evaluate the overall performance of the searched 3D-ADNAS model with the optimal multimodal fusion topology.  Specifically, our proposed model is trained and evaluated according to the settings in \textbf{Preliminaries}.

\section{Evaluating the Improved Fusion Architecture}
We then conduct empirical comparison experiments to validate the effectiveness of 3D-ADNAS.

\subsection{Better 3D-AD performance}
\subsubsection{Benchmark Dataset.} We evaluate 3D-ADNAS on Eyecandies \cite{bonfiglioli2022eyecandies} and MVTec 3D-AD \cite{DBLP:conf/visapp/BergmannJSS22} datasets. Both contain 10 data categories, where MVTec 3D-AD is the real data collected from realistic scenes and Eyecandies is the synthetic virtual data.

\subsubsection{Setup.}
We compare 3D-ADNAS with several baselines (AST \cite{asymmetric}, M3DM \cite{wang2023multimodal}, EasyNet \cite{chen2023easynet}, CFM \cite{zino2024multimodal}, BTF \cite{horwitz2023back}, and Shape\_Guide \cite{chu2023shape}) on both dataset. To make this comparison, we first reset the input image size to $256\times256$ adhering to the settings of \cite{chen2023easynet} and then train the 3D-ADNAS with the Adam optimizer. 

\subsubsection{Evaluations on Eyecandies.} Tables 1 and 2 show the comparison results between  3D-ADNAS and state-of-the-art (SOTA) methods on Eyecandies dataset. It is clear that 3D-ADNAS consistently outperforms the baselines in most test cases. Specifically, when both RGB and 3D depth images are used for training, 3D-ADNAS achieves 4.9\% and 1.6\% higher than M3DM in terms of I-AUROC and AUPRO metrics, respectively (16.6\% and 15.4\% higher than AST, and 6.5\% and 1.1\% higher than CFM, respectively). In fact, the above improvement surpasses the SOTA approaches. These experimental results validate the critical role of multimodal fusion architectures in improving the 3D-AD  performance.

\subsubsection{Evaluations on MVTec 3D-AD.} Table 3 reports the I-AUROC scores of 3D-ADNAS on MVTec 3D-AD dataset. As shown, we can see that 3D-ADNAS obtains best detection results in terms of both RGB-only and multimodal images in most test instances. In particular, 3D-ADNAS improves the I-AUROC value by 0.6\% with 25 times less memory usage than M3DM, and by 1.4\% with approximately 5 times faster frame rate than AST (while using comparable memory usage) (see Table 5). The above results and the visualization of Fig. 5 again demonstrate the effectiveness of our proposed multimodal fusion architecture design in enhancing 3D-AD. Due to the limitation of pages, more experimental results are provided in the Supplementary Material.

\begin{figure}[t]
 \centering
      \includegraphics[scale=0.325]{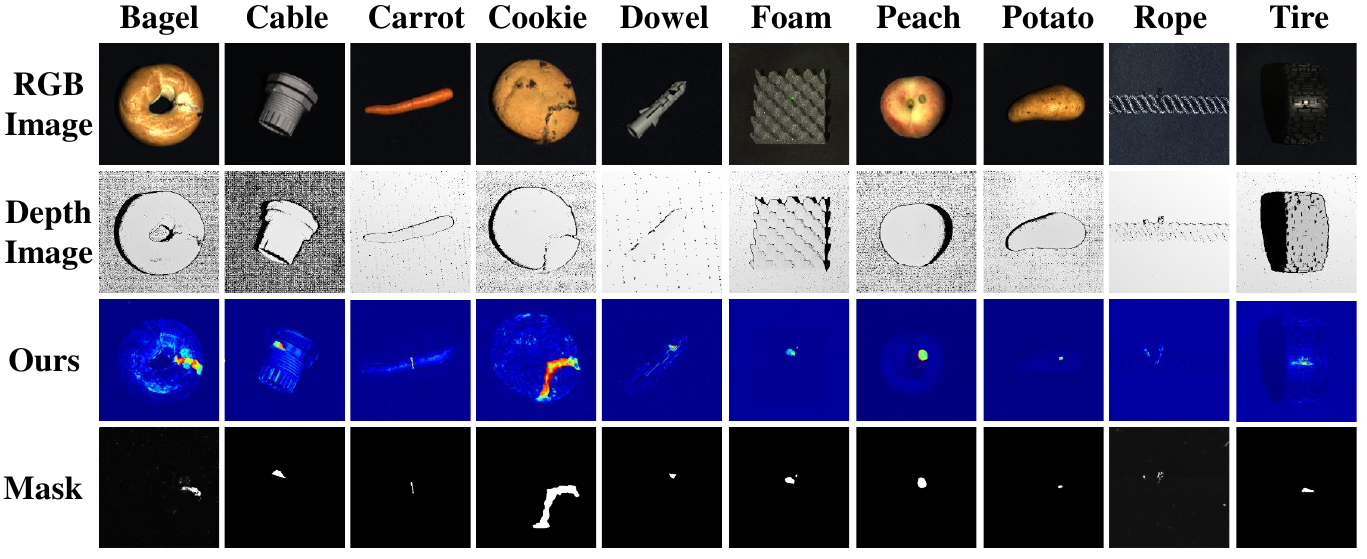}
  \caption{Visualizations results on MVTec 3D-AD.}
 \label{fig:5}
\end{figure}
\subsection{Impact of MSM Fusion Components} 
In this subsection, we again scrutinize the impact of the components of the multimodal fusion architecture design on 3D-AD tasks. As shown in Table 4,  First, we can observe that in all test cases, 3D-ADNAS obtains better results when combining three MSM modules than only exploiting one MSM module, e.g., a 9.8\% improvement over the early MSM, and a 3.0\% improvement over the middle MSM. Second, the performance of 3D-AD can be improved when combining the middle MSM with other MSMs, e.g., when early MSM is combined with middle MSM, the performance is improved by 7.7\% in terms of I-AUROC. Then, it is evident that our proposed two-level search space is able to reach a competitive multimodal fusion architecture for 3D-AD.

\subsection{Higher Frame Rate and Lower Memory Usage} To further show the advantage of the proposed method, we report the results in terms of memory usage, frame rate, and I-AUROC on MVTec 3D-AD. As shown in Table 5, we can see that 3D-ADNAS gets fastest frame rate, highest I-AUROC scores, and lowest memory usage when tested on single NVIDIA RTX 4090. \textbf{What contributes to the competitive performance of 3D-ADNAS?} Intuitively, it benefits from the fact that our method uses neither memory-bank-based strategies that increase memory nor large pre-trained language models that affect inference speed. In this sense, revisiting multimodal fusion from an architectural perspective is indeed an effective scheme for advancing 3D-AD.

\subsection{Study on Few-shot Anomaly Detection} In resource-constrained scenarios, collecting a large number of samples is extremely expensive and infeasible. Thus, few-shot anomaly detection \cite{kim2024few,duan2023few} becomes a promising solution. To evaluate the effectiveness of 3D-ADNAS in few-shot scenarios, we randomly select 5, 10, and 50 images on Eyecandies and MVTec 3D-AD datasets as the training set and perform inference on the full test set. As shown in Tables 6 and 7, 3D-ADNAS still exhibits promising performance.

\section{Related Work}
\subsubsection{3D Anomaly Detection.} Existing studies mainly focus on design of 2D-AD methods, i.e., detecting flaws in RGB images \cite{DBLP:BergmannFSS19, jiang2022softpatch, hu2024anomalydiffusion}, including feature-embedding-based methods \cite{liu2023simplenet,li2021cutpaste,rudolph2022fully, rudolph2021same, deng2022anomaly, bergmann2020uninformed,cohen2020sub, defard2021padim} and reconstruction-based methods \cite{schluter2022natural,zavrtanik2021draem,you2022unified,zavrtanik2022dsr}. Recently, 3D-AD has emerged as an improvement of 2D-3D, and received a surge of attention \cite{liu2024deep,xie2024iad}, since it exhibits a powerful detection ability via exploiting both RGB and depth images rather than only RGB \cite{reiss2022anomaly,zavrtanik2024keep,zhao2024pointcore}. A series of 3D-AD methods have been proposed and developed, e.g., BTF \cite{horwitz2023back}, M3DM \cite{wang2023multimodal}, and CFM \cite{zino2024multimodal}. However, existing works rarely focus on the impact of architectural design on 3D-AD. To bridge a gap, we systematically study the impact of multimodal fusion architecture design via theoretical analysis and experiments, and then propose a simple yet effective method.

\textbf{Neural Architecture Search.} NAS aims to automate the design of task-specific deep neural network architectures, which can be formulated as an optimization problem \cite{DBLP:conf/iclr/BakerGNR17,zoph2018learning}.  In this process, NAS uses a search strategy to traverse a specified search space comprising candidate architectures. Then, the architecture with the best performance is selected as the final design. It has been empirically demonstrated that NAS can shape architectures that surpass those manually designed \cite{yu2020deep, lv2024multiscale}. Note that, the key of applying NAS to 3D-AD lies in defining a suitable search space, which depicts a searchable subset of candidate architectures from the vast architecture space. Given the limited prior knowledge regarding the impact of architectural designs on 3D-AD tasks, defining a suitable search space remains challenging.

\section{Conclusion}
 This paper investigates the impact of two-level multimodal fusion architecture design (including intra-module and inter-module fusion levels) on 3D-AD tasks. The proposed 3D-ADNAS bridges the gap between multimodal fusion architecture design and 3D-AD, achieving comprehensive improvement of 3D-AD performance in terms of detection accuracy, frame rate and memory usage. We believe that devising a friendly multimodal fusion architecture is practically meaningful for 3D-AD, and hope this work inspires further research on 3D-AD from the architectural perspective. 

\section{Acknowledgments}
This work is supported by the National Natural Science Foundation of China under Grant 62472079.

\bibliography{aaai25}

\end{document}